\documentclass[11pt]{article}

\usepackage[preprint]{acl}

\usepackage{times}
\usepackage{latexsym}

\usepackage[T1]{fontenc}

\usepackage[utf8]{inputenc}

\usepackage{microtype}

\usepackage{inconsolata}

\usepackage{graphicx}

\usepackage{amsmath,amssymb,amsfonts}
\usepackage{booktabs}
\usepackage{float}
\usepackage{caption}
\usepackage{multicol,multirow}
\usepackage{tcolorbox}
\usepackage[section]{placeins}
\usepackage{makecell}
\usepackage{enumitem}
\usepackage{marvosym}

\title{SaRoHead: Detecting Satire in a Multi-Domain Romanian News Headline Dataset}

\author{
\textbf{Mihnea-Alexandru Vîrlan\textsuperscript{1}},
\textbf{Răzvan-Alexandru Smădu\textsuperscript{1}},
\textbf{Dumitru-Clementin Cercel\textsuperscript{1,}}\thanks{Corresponding Author},\\
\textbf{Florin Pop\textsuperscript{1,2}},
\textbf{Mihaela-Claudia Cercel\textsuperscript{3,4}}
\\
\textsuperscript{1}National University of Science and Technology POLITEHNICA Bucharest,\\
Faculty of Automatic Control and Computers, Bucharest, Romania\\
\textsuperscript{2}National Institute for Research \& Development in Informatics - ICI Bucharest, \\
Bucharest, Romania \\
\textsuperscript{3}Paris 1 Panthéon-Sorbonne University, Paris, France\\
\textsuperscript{4}University of Bucharest, Bucharest, Romania\\
\texttt{dumitru.cercel@upb.ro}
}

\begin{document}

\maketitle

\begin{abstract}
The primary goal of a news headline is to summarize an event in as few words as possible. Depending on the media outlet, a headline can serve as a means to objectively deliver a summary or improve its visibility. For the latter, specific publications may employ stylistic approaches that incorporate the use of sarcasm, irony, and exaggeration, key elements of a satirical approach. As such, even the headline must reflect the tone of the satirical main content. Current approaches for the Romanian language tend to detect the non-conventional tone (i.e., satire and clickbait) of the news content by combining both the main article and the headline. Because we consider a headline to be merely a brief summary of the main article, we investigate in this paper the presence of satirical tone in headlines alone, testing multiple baselines ranging from standard machine learning algorithms to deep learning models. Our experiments show that Bidirectional Transformer models outperform both standard machine-learning approaches and Large Language Models (LLMs), particularly when the meta-learning Reptile approach is employed. 
\end{abstract}

\section{Introduction}

The complex nature of satire makes it difficult to define it \cite{zekavat2014discursive,condren2012satire}, but it is commonly agreed that satire represents a way to ridicule and criticize certain aspects of society. As such, it can also be used to present an alternative, humorous version of some events covered in regular news outlets, bringing a subtle and refined criticism. This subtility enhances ambiguity \cite{landreville2015satire}, which makes satirical content more challenging to understand without appropriate context.

Since the primary goal of a news headline is to give a summary, the way it is presented influences the decision about whether or not to read the main article. \citet{DBLP:conf/icwsm/ParkKAC21} conducted a study on regular and edited headlines coming from media outlets, and found that significant changes in the social media headline lead to lower similarity compared to the original headline. Additionally, they found that the edited headlines tend to have a clickbait-inclined form, designed to capture the viewers' attention. Given this format, clickbait could be behind fake news 
and as such, \citet{das2019satire} showed that a satirical headline is more likely to express feelings, in terms of tonality.

Detecting forms of humor in Natural Language Processing is a topic of interest, as numerous approaches have considered the usage of classic machine learning algorithms \citep{DBLP:conf/wanlp/NayelAAA21}, ensemble methods \citep{DBLP:conf/acl-figlang/LemmensBLMD20,DBLP:conf/aclnut/BadlaniAR19}, and modern deep learning \citep{DBLP:conf/acl-figlang/GregoryLMTDR20,jang2024context,DBLP:conf/acl-figlang/AvvaruVM20}.

To extend existing work on satire detection in Romanian \citep{DBLP:conf/nldb/EchimSACP23,DBLP:conf/acl/RogozGI20}, we propose a new dataset, namely \textbf{SaRoHead}, made from headlines gathered from Romanian news outlets, as well as a known online publication for satirical content. Each of these belongs to a specific news category that groups related headlines. On this dataset, we evaluate baselines ranging from classic machine learning algorithms, BERT \cite{devlin2019bert} models and their variations, and Large Language Models (LLMs). For the BERT models, we investigate the impact of intermediate unsupervised training on a related dataset and the usage of Reptile \citep{DBLP:journals/corr/abs-1803-02999}. For the LLMs, we compare various few-shot settings with LoRA \citep{DBLP:journals/corr/abs-2106-09685}. Since we believe each news category has its own nuances, we experiment with each category separately.

This work brings the following contributions:
\begin{itemize}
    \item We propose a dataset consisting of news headlines gathered from various online news outlets and a satirical outlet. We group this news by the overall category it represents. The dataset will be made public.
    \item We perform experiments on various classification approaches, by training each model category-wise. The proposed baselines encompass classic machine learning algorithms, BERT models, and large language models.
    \item We evaluate three variations of BERT using the intermediate task transfer learning approach by identifying which intermediate task yields the best results. For the intermediate task, we chose datasets that we deem similar to the one we introduced. For comparison, we also apply the meta-learning Reptile algorithm. 
\end{itemize}

\section{Related Work}

\subsection{Detection of Non-conventional Styles}

\citet{DBLP:journals/jwe/KamalAJ24} used a two-layer BiLSTM and self-attention to detect satirical content in short texts by appending handcrafted features to the outputs of the second BiLSTM layer. \citet{DBLP:journals/access/RazaliHCND22} compared machine learning classifiers on five sets of features for satire detection: one was extracted using a CNN from n-grams, while the rest represented linguistic features. 
Another approach \cite{ahuja2022transformer} combined a recurrent CNN with RoBERTa \citep{liu2019robertarobustlyoptimizedbert} into a hybrid architecture.
\citet{pandey2023bert} utilized BERT together with an LSTM on top, and trained the model on a language mix of Indian and English. The results showed that the features used in training the classical machine learning algorithms are not as rich as those of the hybrid model. 

\citet{wang2024cross} employed a multi-modal approach to detect sarcasm in which the standard multi-head self-attention mechanism was modified such that a query representation of an image attends to relevant text parts and vice-versa. This helps the model catch relevant patterns and ensures a clear and mutual influence of the input text and image. 

An approach based on Large Language Models \cite{yao2024sarcasm} detects sarcasm by using the chain of thought to find smaller subproblems. More concretely, the authors proposed the chain of contradiction, the graph of cues, and the tensor of cues, which outperform existing approaches. 
Other works have evaluated manually extracted features \cite{razali2021sarcasm}, as well as multi-modal approaches incorporating audio \cite{bedi2021multi} and images \cite{pan2020modeling}.

\subsection{Impact of Feature Selection}

\citet{georgieva2021research} used a variation of the mutual information to consider the number word frequency contained in documents belonging to a specific class, which improved the results when compared to the standard mutual information. \citet{alassaf2022improving} employed ANOVA (analysis of variance) on a dataset of tweets in the Arabic language, and applied stemming to reduce the vocabulary. Multiple classical algorithms, such as SVM, Naive Bayes, and linear layers, were evaluated
and the results significantly improved, while those of the others remained either similar or decreased.

\subsection{Intermediate Task Transfer Learning}

Prior work \cite{pruksachatkun-etal-2020-intermediate,nkhata2025intermediatetasktransferlearningleveraging,park-caragea-2020-scientific,savini2022intermediate} aimed at pretraining a BERT-like model on a similar task and evaluating the performance with respect to the target task. The results can vary depending on the complexity of the intermediate task and the intrinsic linguistic features.
\citet{sosea2023sarcasm} proposed to predict whether a Tweet that follows a disaster, such as a hurricane, is sarcastic or not. For the intermediate tasks, they identified datasets that address emotion detection, but also other varieties of sarcasm, such as iSarcasm \cite{oprea2019isarcasm}, and SARC \cite{khodak2018largeselfannotatedcorpussarcasm}. 
\citet{sosea2020canceremo} investigated the usage of intermediate training using the BERT language model when predicting emotions in diagnosed patients with cancer. It showed that results depend on similarities between the intermediate task and the target task. 

\begin{table*}[!t]
    \centering
    {
    \scriptsize
    \begin{tabular}{lcccc}
         \toprule
         \textbf{Dataset} & \textbf{Num. samples}  &
         \textbf{Domain} &
         \textbf{Content} & \textbf{Language}\\
         \midrule
         SciTechBaitRo \cite{ginga-uban-2024-scitechbaitro} & 10,640 & Science\&Technology & Titles only & Romanian\\
         SaRoCo \cite{DBLP:conf/acl/RogozGI20}& 55,608 &  N/A & Title or/and article & Romanian \\
         RoCliCo \cite{broscoteanu2023novelcontrastivelearningmethod} & 8,313 & N/A & Title and article & Romanian\\
         FreSaDa \cite{DBLP:conf/ijcnn/IonescuC21} &11,570 & N/A & Title and article & French\\
         News Headlines Dataset \cite{DBLP:journals/corr/abs-1908-07414} & 26,709 & N/A & Titles only & English\\
        \midrule 
         SaRoHead (ours) & 20,725 & Social, politics, sports & Titles only&Romanian \\
         \bottomrule
    \end{tabular}
    }
    \caption{Our dataset compared with other existing similar datasets.}
    \label{dataset_comp}
\end{table*}

Intermediate task transfer learning has also been used in keyphrase identification and classification \citep{park2020scientific}.
The authors explored several intermediate tasks, including assigning word labels based on semantic and grammatical values, as well as tasks of Natural Language Inference. Since two variants of BERT were also considered, namely the standard one \citep{devlin2019bert} and a variant pre-trained on scientific text \citep{beltagy2019scibert}, the results showed that SciBERT achieved significant improvements using the intermediate tasks, primarily due to the similarities with the data it was pre-trained on.

\section{Dataset}

\subsection{Data Collection}

We scraped the news headlines from various satirical and non-satirical news outlets. We used the site maps available on each platform, and the Python package BeautifulSoup\footnote{\url{https://beautiful-soup-4.readthedocs.io/en/latest/}} to extract the titles.
The satirical content was collected from TimesNewRoman\footnote{\url{https://www.timesnewroman.ro/}}, while the regular, non-satirical headlines were collected from DCNews\footnote{\url{https://www.dcnews.ro/}}, Antena 3\footnote{\url{https://www.antena3.ro/}}, Mediafax\footnote{\url{https://www.mediafax.ro/}} and sport.ro\footnote{\url{https://www.sport.ro/}}.
The following news domains were selected to enrich the dataset's diversity: \textit{social}, \textit{politics}, and \textit{sports}.
While gathering the data, we noticed that the URL links from TimesNewRoman, Antena 3, and Mediafax explicitly contain the keywords for categories, whereas those from DCNews do not. As a result, we looked for keywords relevant to each domain. For regular sports news headlines, we used the sport.ro news outlet.
The corpus is diverse, with headlines spanning the period from 2009 to 2025, with a cutoff date of June 2025. Ultimately, our dataset comprises 20,725 news headlines from publicly available Romanian news outlets. 
We call this dataset SaRoHead (\textbf{Sa}tirical \textbf{Ro}manian \textbf{Head}lines). Table \ref{dataset_comp} provides a comparison between our proposed dataset and similar datasets in Romanian, English, and French.

\subsection{Data Labelling}

\begin{figure}[!t]
    \centering
    \includegraphics[width=0.95\columnwidth]{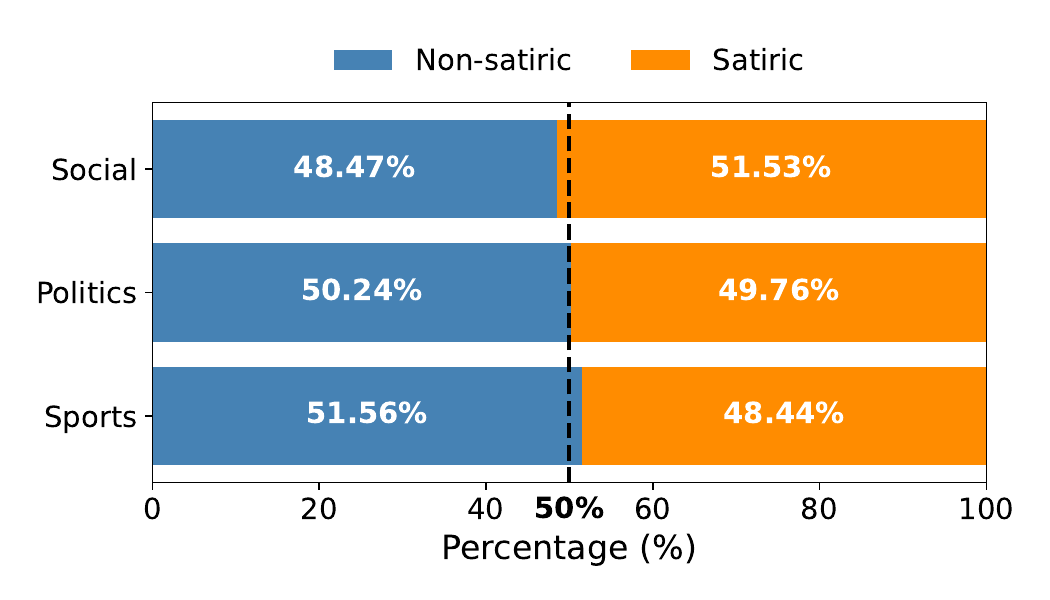}
    \caption{Balancing of the per-domain label distribution of SaRoHead.}
    \label{satire_domain_distrib}
\end{figure}

For labeling the data, similar to \citet{kiesel-etal-2019-semeval}, we did not rely on human labelers but instead used distant supervision at the source news outlet level. As such, since TimesNewRoman is a well-known satirical outlet, we considered the ground truth of their headlines to be satirical. For headlines coming from regular, mainstream sources, we labeled them as non-satirical.
The balance in distributions is showcased across the news categories in Figure \ref{satire_domain_distrib}.

\subsection{Data Processing}

When processing the dataset, we took the following steps.

\textbf{Removal of diacritics.} To normalize the satire detection dataset, we decided to remove the diacritics and replace them with English equivalent letters. 

\textbf{Named entity replacement.} Since the headlines often contain names, personalities, geographic locations, and facilities, we do not want to bias the results of the models. Moreover, this could pose some ethical challenges. As such, similar to \citet{rogoz-etal-2021-saroco}, we replace the named entities with placeholders using Spacy's module for the Romanian language\footnote{\url{https://spacy.io/models/ro}}. The considered placeholders for entities are the following: \texttt{[FACILITY]}, \texttt{[NAT\_REL\_POL]}, \texttt{[ORGANIZATION]},
\texttt{[PERSON]}, and \texttt{[ORGANIZATION]}. To account for the imperfect nature of the NER module, manual replacements were enforced for certain words.

\begin{table}[t]
    \centering
    {
    \small
    \begin{tabular}{lrr}
        \toprule
        \textbf{Category} &  \textbf{Non-satiric} &  \textbf{Satiric} \\
        \midrule
        Social   &                 4,770 &             5,071 \\
        Politic  &                 4,434 &             4,391 \\
        Sport    &                 1,072 &             1,007 \\
        \midrule
        Total    &                10,276 &            10,469 \\
        \bottomrule
        \end{tabular}
    }
    \caption{Number of non-satiric and satiric samples}
    \label{raw_samples}
\end{table}

\begin{table}[t]
\centering
{
\small
\begin{tabular}{lrrrlllll}
\toprule
\textbf{News Topic} &  \textbf{Training} &  \textbf{Validation} &  \textbf{Test} \\
\midrule
    Social &        7,225 &             1,963 &        653 \\
   Politics &        6,719 &             1,579 &        527 \\
     Sport &        1,576 &              378 &        125 & \\
\bottomrule
\end{tabular}
}
\caption{Dataset statistics}
\label{data_stats}
\end{table}

\subsection{Dataset Statistics}

Figure \ref{satire_domain_distrib} and Table \ref{raw_samples} show the number of samples per category, indicating the balanced nature. In total, each label contains approximately 10,000 samples, with social and political labels having roughly the same number of samples (i.e., 4,000-5,000). In comparison, the sports label has 1,000 samples per label, which underrepresents the category. The dataset used in the experiments was split into training, validation, and test sets. Table \ref{data_stats} presents the raw number of samples for each of the three splits.

\section{Experiments}

Our goal is to evaluate various baselines for detecting Romanian satire in the headlines and assess their performance. We compare the following approaches: \textbf{standard machine learning}, \textbf{BERT-based approaches}, and \textbf{large language models}.

\subsection{Standard Machine Learning Algorithms}

We utilize standard algorithms in conjunction with feature selection techniques to assess whether their performance is dependent on the number of relevant features.

\subsubsection{Algorithms}

We explore the usage of several machine learning classifiers, with implementations available in scikit-learn \cite{DBLP:journals/jmlr/PedregosaVGMTGBPWDVPCBPD11}:
Support Vector Machines (SVM) \cite{cortes1995support}, Random Forest \cite{DBLP:journals/ml/Breiman01}, Logistic Regression, and Naive Bayes.

We opt to use the default hyperparameters as follows:
\textbf{SVM:} margin is set to 1.0, and employs the radial basis function (RBF) kernel \cite{DBLP:journals/compsys/BroomheadL88};
\textbf{Random Forest:} we use the Gini entropy \cite{breimanclassification};
\textbf{Naive Bayes:} we set the Laplace smoothing factor $\alpha=1.0$;
\textbf{Logistic Regression:} we set the regularization parameter $C=1.0$.

\subsubsection{Features and Feature Selection}

The features represent the set of all unigrams, bi-grams, and tri-grams, extracted with the help of CountVectorizer, implemented in scikit-learn\footnote{\url{https://scikit-learn.org/stable/modules/generated/sklearn.feature_extraction.text.CountVectorizer.html}}.
We consider the following feature selection techniques: 

\textbf{Mutual Information.} The Mutual Information (MI) \cite{DBLP:journals/bstj/Shannon48} measures the importance of a word given a label. It can be used before inputting the preprocessed data in neural networks \cite{abdulghani2022hybrid} to filter out irrelevant terms. 

\textbf{Chi-Squared test.} The Chi-Squared (CHI2) statistical test \cite{DBLP:conf/ictai/LiuS95} determines the link between two features by summing the squared difference of the observed and expected values.

\textbf{ANOVA.} Another statistical test, analysis of variance (ANOVA) \cite{fisher1970statistical}, builds upon two types of variance: between groups and within groups. 

Each of these techniques is used to select a number $F$ of relevant features. The following values for $F$ are considered: $F\in \{500,1000,1500,3000,5000,7000\}$.

\subsection{BERT-based Methods}

We employ several BERT-based \cite{devlin2019bert} models in the unsupervised task transfer learning and meta-learning setting using Reptile \cite{DBLP:journals/corr/abs-1803-02999}.

\subsubsection{Unsupervised Task Transfer Learning}

In this set of experiments, we not only directly fine-tune the models on our downstream task, but also explore the influence of intermediate task transfer learning. That means we first train the model on the intermediate task, then we continue the training on the downstream task.

We train three Transformer-based neural networks: a Romanian pretrained BERT \cite{DBLP:journals/corr/abs-2009-08712}, XLM-RoBERTa \cite{conneau2020unsupervisedcrosslingualrepresentationlearning}, and a Romanian pretrained Distilled-BERT \cite{avram-etal-2022-distilling}. XLM-RoBERTa is a version of RoBERTa \cite{liu2019robertarobustlyoptimizedbert} trained on 100 languages, including Romanian.

For the intermediate task transfer learning, we train every model on all layers using the unsupervised task of masked language modeling. We use a linear learning rate scheduler, with the initial value of \(10^{-5}\) and the following intermediate datasets: SaRoCo \cite{DBLP:conf/acl/RogozGI20} for 40 epochs, SciTechBaitRo \cite{ginga-uban-2024-scitechbaitro} for 45 epochs, and RoCliCo \cite{broscoteanu2023novelcontrastivelearningmethod} for 50 epochs. For each one of them, we work only with the headlines.

For all our experiments, when fine-tuning on the target task (in our case, SaRoHead), we do it separately on every news category. We only train a linear layer that takes as input the CLS embedding token while keeping the backbone frozen. We train for 35 epochs using the Adam optimizer, with a batch size of 32. Across each batch, we pad the input to the maximum length of a tokenized text. Since we use the linear learning rate scheduler, we perform a grid search on the starting learning rate from the set \(\{10^{-3},10^{-4},10^{-5}\}\) and we take the value of \(10^{-3}\).

\subsubsection{Reptile}

We also explore the usage of the Reptile meta-learning technique \cite{DBLP:journals/corr/abs-1803-02999}. It is a first-order meta-learning algorithm that trains the model to quickly adapt to new tasks with only a few gradient steps. It involves sampling tasks from a pool, and for each task, we make several updates, using the Adam \cite{kingma2017adammethodstochasticoptimization} optimizer. We ensure that for each task, we maintain separate gradients. Only after the training step on the task pool is complete, we update the original parameters.
The hyperparameters are the following: \(K\) represents the number of examples sampled from a task; $\alpha$ is a parameter used during the meta-learning step; $\beta$ is the learning rate for the Adam optimizer.

To find the hyperparameters, we use grid search on \(K\in \{5,7\}\), \(\beta \in \{10^{-4},10^{-3},10^{-2}\}\), \(\alpha \in \{0.1,0.3,0.5,0.8\}\) and keep the batch size set to 32. 
The reported results are using \(K=7,\beta=10^{-4},\alpha=0.8\).
The update rule when computing task-specific parameters is the Adam optimizer, and the number of meta epochs is set to 150. We also use a linear decrease in $\alpha$ during training. The tasks used in Reptile are identical to those employed in the unsupervised task transfer setup. When doing the meta-learning, we do it for every news topic. As such, the tasks contain not only the three intermediate datasets, but also the subset for which we need to train using Reptile. A difference is that we update all the parameter models.

\subsection{Large Language Models}

We explore the effectiveness of Large Language Models by experimenting with both zero- and few-shot learning-based classification, as well as fine-tuning. We use four Romanian instruction-tuned LLMs \cite{masala2024vorbecstiromanecsterecipetrain}, namely: RoLlama2 based on Llama2 7B \cite{DBLP:journals/corr/abs-2307-09288}, RoLlama3 based on Llama3 8B \cite{DBLP:journals/corr/abs-2407-21783}, RoGemma2 based on Gemma 2 9B \cite{DBLP:journals/corr/abs-2403-08295}, and RoMistral based on Mistral 7B \cite{DBLP:journals/corr/abs-2310-06825}.

\subsubsection{Zero- and Few-shot Learning}

For the zero- and few-shot setups, we utilize a prompt that asks an LLM to classify the headline, also taking into account the labeled entities. The prompts in Romanian and their English translations are presented in Appendix \ref{app:prompts}.

For each model and news topic, we vary the number of examples provided to the LLM, \(n\in \{0,2,4,6\}\). If \(n\neq 0\), then the model will receive \(\frac{n}{2}\) (i.e., half) satirical examples and \(\frac{n}{2}\) (i.e., half) non-satirical examples. 

\subsubsection{LoRA Fine-tuning}

Low-rank Adaptation (LoRA) \cite{DBLP:journals/corr/abs-2106-09685} is an LLM fine-tuning technique in which, given the set of parameters \(W\), it does not update them directly, due to compute constraints, but rather directly finds and updates an update \(\Delta W\approx A\cdot B\), where if \(\Delta W\in \mathbf{R}^{n\times m}\), then \(A\in \mathbf{R}^{n\times r}\) and \(B\in \mathbf{R}^{r\times m}\). During the update $W_{adapted} = W + \frac{\alpha}{r} \Delta W $, it uses a scaling factor, namely $\alpha$, and a rank $r$ of the low-rank matrices.

Prior work on this technique has demonstrated a performance boost when applied to classify synthetic data that it previously generated \cite{DBLP:journals/corr/abs-2407-19299} under data scarcity conditions. Also, varying the hyperparameters $\alpha$ and \(r\) has been shown to influence the behaviour of the LLM \cite{app15063087,r-etal-2024-shot}. \citet{DBLP:journals/kbs/GaoMLSJJWNYXCY24} has demonstrated that multiple LoRA adapters trained on different tasks can be fused into the backbone LLM to improve the capability of following human instructions. 

We use grid search on the parameters $\alpha$ and $r$ for the following values $\alpha \in \{8,16\}$ and $r \in \{4,8,16\}$ and the number of epochs $N \in \{3,5,7\}$. From there, we set $\alpha=8$ , \(r=8\), the dropout to 0.01 and $N$ to 5. The optimizer used is Adam with a weight decay of \(10^{-2}\). We fine-tune the \(W_Q,W_K,W_V, W_o\) matrices from the self-attention module and set the device batch size to 8.

\subsection{Experiments Evaluation}

The reported experiments are run and evaluated on every domain from the dataset, individually. We report the average of three results (i.e., for social, politics, and sports) over a single training run.

We evaluate each class of models by the following metrics, only on the satiric-label:
\textbf{Precision (P)} scores how many of the samples that were predicted as positive match the ground truth;
\textbf{Recall (R)} scores how many of the total positive samples were correctly predicted as positive;
\textbf{F1-score (F1)} is the harmonic mean between the precision and recall.

\section{Results}

\subsection{Standard Machine Learning Algorithms}

Table \ref{T3} reports the average metrics obtained for each algorithm and the top number of features across all feature selection techniques and news topics for the test set. For all the algorithms involved, the peak performance was achieved when using 1,500 features, and subsequently, the F1-scores decreased to the point where those for regular headline detection were higher than those for their counterparts in the other class.

In Table \ref{T5}, we show that, on average, regardless of the used technique, the results are similar. The result that remained consistent throughout the validation and test splits was the use of ANOVA for Logistic Regression, which maintained the highest F1-score for non-satire content detection.

When it comes to the effect of an algorithm on a news topic, Table \ref{T7} suggests that SVM classifies satirical content from the political spectrum with confidence. It also shows that, topic-wise, SVM, together with Logistic Regression, are the two best algorithms for satire detection, and Random Forest achieves the lowest F1-score for non-satirical content detection.

\begin{table}[ht]
    \centering
{
\small
\begin{tabular}{lclll}
\toprule
 \textbf{Algorithm} & \textbf{\# Features} & \textbf{R (\%)} & \textbf{P (\%)} & \textbf{F1 (\%)} \\
\midrule
\multirow[c]{6}{*}{SVM} & 500 & 82.41 & 71.25 & 76.21 \\
 & 1,000 & 83.76 & 74.46 & 78.55 \\
 & 1,500 & \textbf{\underline{87.63}} & 73.91 & \textbf{\underline{79.91}} \\
 & 3,000 & 75.75 & 79.22 & 75.43 \\
 & 5,000 & 66.05 & \underline{81.02} & 64.58 \\
 & 7,000 & 61.86 & 74.67 & 57.84 \\
\midrule
\multirow[c]{6}{*}{RF} & 500 & 80.91 & 69.41 & 74.56 \\
 & 1,000 & 76.82 & 71.75 & 73.42 \\
 & 1,500 & \underline{83.13} & 71.68 & \underline{76.81} \\
 & 3,000 & 74.72 & 72.22 & 72.46 \\
 & 5,000 & 67.43 & \underline{74.69} & 65.33 \\
 & 7,000 & 64.02 & 73.13 & 61.21 \\
\midrule
\multirow[c]{6}{*}{NB} & 500 & 69.05 & 74.37 & 70.76 \\
 & 1,000 & 70.11 & 76.90 & 72.97 \\
 & 1,500 & \underline{75.26} & 78.21 & \underline{76.59} \\
 & 3,000 & 67.43 & 80.29 & 72.93 \\
 & 5,000 & 68.75 & 81.60 & 73.59 \\
 & 7,000 & 64.42 & \textbf{\underline{84.79}} & 71.72 \\
\midrule
\multirow[c]{6}{*}{LR} & 500 & 81.28 & 73.07 & 76.87 \\
 & 1,000 & 81.33 & 76.78 & 78.81 \\
 & 1,500 & \underline{82.68} & 77.41 & \underline{79.81} \\
 & 3,000 & 81.11 & 77.76 & 79.21 \\
 & 5,000 & 76.34 & 78.83 & 76.49 \\
 & 7,000 & 72.80 & \underline{79.59} & 74.56 \\
\bottomrule
\end{tabular}
}
\caption{Average metrics for each algorithm and top number of features on the testing set. Underline indicates the best score for each algorithm, and bold indicates the best overall scores.}
\label{T3}
\end{table}

\begin{table}[!t]
    \centering
{
\small
\begin{tabular}{lllll}
\toprule
\textbf{Algorithm} & \makecell{\textbf{Feature}\\ \textbf{Selection}} & \textbf{R (\%)} & \textbf{P (\%)} & \textbf{F1 (\%)} \\
\midrule
\multirow[c]{3}{*}{SVM} & ANOVA & 76.18 & \underline{75.87} & \underline{72.13} \\
 & CHI2 & 76.17 & 75.78 & 72.08 \\
 & MI & \underline{76.38} & 75.62 & 72.06 \\
\midrule
\multirow[c]{3}{*}{RF} & ANOVA & 73.81 & 72.14 & 70.04 \\
 & CHI2 & 73.95 & 72.07 & 70.16 \\
 & MI & \underline{75.76} & \underline{72.23} & \underline{71.69} \\
\midrule
\multirow[c]{3}{*}{NB} & ANOVA & 69.01 & 79.54 & 73.10 \\
 & CHI2 & \underline{69.38} & \textbf{\underline{79.66}} & \underline{73.39} \\
 & MI & 69.12 & 78.88 & 72.79 \\
\midrule
\multirow[c]{3}{*}{LR} & ANOVA & 79.18 & 77.23 & 77.60 \\
 & CHI2 & 79.11 & \underline{77.26} & 77.58 \\
 & MI & \textbf{\underline{79.48}} & 77.23 & \textbf{\underline{77.70}} \\
\bottomrule
\end{tabular}
}
    \caption{Average metrics for each algorithm and each feature selection technique for the testing set. Underline indicates the best score for each algorithm, and bold indicates the best overall scores.}
    \label{T5}
\end{table}

\begin{table}[!t]
{
    \centering
    \small
\begin{tabular}{lllll}
\toprule
\textbf{News Topic} & \textbf{Algorithm} & \textbf{R (\%)} & \textbf{P (\%)} & \textbf{F1 (\%)} \\
\midrule
\multirow[c]{4}{*}{Social} & SVM & \textbf{\underline{88.97}} & 66.52 & 76.12 \\
 & RF & 86.53 & 62.62 & 72.62 \\
 & NB & 76.60 & 71.32 & 73.76 \\
 & LR & 84.70 & \underline{69.54} & \underline{76.36} \\
\midrule
\multirow[c]{4}{*}{Politic} & SVM & \underline{88.66} & 86.37 & \textbf{\underline{87.49}} \\
 & RF & 86.54 & 81.23 & 83.77 \\
 & NB & 75.17 & \textbf{\underline{87.17}} & 80.71 \\
 & LR & 86.13 & 87.12 & 86.61 \\
\midrule
\multirow[c]{4}{*}{Sport} & SVM & 51.09 & 74.37 & 52.66 \\
 & RF & 50.46 & 72.59 & 55.51 \\
 & NB & 55.74 & \underline{79.59} & 64.81 \\
 & LR & \underline{66.94} & 75.07 & \underline{69.90} \\
\bottomrule
\end{tabular}
}
    \caption{Average metrics for each news topic and algorithm for the testing set. Underline indicates the best score on each topic, and bold indicates the best overall scores.}
    \label{T7}
\end{table}

\subsection{BERT-based Methods}

\subsubsection{Intermediate Unsupervised Task Transfer Learning}

Table \ref{tabel_average_test_BERT} summarizes the average results obtained for every model across the news topics for the test set. Of all the models tested, BERT achieved the best F1-scores, alongside all the other models. Notably, RoCliCo and SciTechBaitRo datasets helped BERT in correctly classifying more examples with a clickbait writing style. SaRoCo also improved the standard BERT model, although the improvement on the F1-scores is not as significant as with the other two intermediate tasks.
For Distilled-BERT, we observe that the intermediate tasks caused confusion, resulting in lower F1-scores. As a result, this lighter model lacks an understanding of some subtleties that differentiate the satirical tone from a regular tone.
When examining XLM-RoBERTa's scores, we observe that they are lower than those of the other models. When compared to the other two models, which were pre-trained on Romanian-only text, the intermediate tasks cannot replace the model's lack of specific linguistic patterns in a given language, as this model has been trained to catch general patterns.

Table \ref{avg_news_topic_test} presents the average metrics across the models for each news topic in the test set. The computation of average metrics is as follows. For a given news topic and model, let \(y_{m,t}^{l}\) be the value of the metric \(m\), unsupervised task \(t\), and the classification label \(l\). Then an entry to the table is $\overline{y_{m}^{l}}=\frac{\sum_{t}y_{m,t}^{l}}{n}$ where \(n\) is the number of intermediate tasks, including the metric value for the standard training that has not been intermediately trained on a task.
Topic-wise results show that BERT achieves the highest F1-scores when compared to the other two models. This highlights the adaptability and flexibility of this neural network, particularly in tasks similar to clickbait detection, as shown in Table \ref{tabel_average_test_BERT}. The highest F1-scores have been obtained on the politics topic. This highlights the adaptability and flexibility of the model, especially when aided by relevant intermediate tasks, as shown in Table \ref{tabel_average_test_BERT}. The results also show that XLM-RoBERTa does not catch specific satire patterns in a given language, explaining why it obtained the lowest scores topic-wise.
We can also see that for any model, the F1-scores for politics surpass those for social and sports, thus indicating that in Romanian headlines, satire is biased towards events on the political spectrum.

\begin{table}[!t]
    \centering
    \small
{\begin{tabular}{llrrr}
\toprule
\makecell{\textbf{News} \\ \textbf{Topic}} & \textbf{Model} & \textbf{R (\%)} & \textbf{P (\%)} & \textbf{F1 (\%)} \\
\midrule
\multirow[c]{3}{*}{Social} & BERT & \textbf{\underline{94.54}} & 74.37 & \underline{83.23} \\
 & Distilled-BERT & 89.00 & \underline{76.71} & 82.40 \\
 & XLM-RoBERTa & 83.16 & 61.96 & 71.01 \\
\midrule
\multirow[c]{3}{*}{Politic} & BERT & \underline{89.34} & \textbf{\underline{93.76}} & \textbf{\underline{91.49}} \\
 & Distilled-BERT & 84.09 & 92.31 & 87.97 \\
 & XLM-RoBERTa & 75.79 & 81.22 & 78.35 \\
\midrule
\multirow[c]{3}{*}{Sport} & BERT & \underline{81.15} & \underline{90.93} & \underline{85.67} \\
 & Distilled-BERT & 78.28 & 82.88 & 80.35 \\
 & XLM-RoBERTa & 61.48 & 74.53 & 66.87 \\
\bottomrule
\end{tabular}
}
    \caption{Test set average metrics across model intermediate tasks. Underline indicates the best score on each topic, and bold indicates the best overall scores.}
    \label{avg_news_topic_test}
\end{table}

\begin{table}[!t]
    \centering
    {
    \scriptsize
\begin{tabular}{lllll}
\toprule
\textbf{Model} & \makecell{\textbf{Intermediate} \\ \textbf{Task}} & \textbf{R (\%)} & \textbf{P (\%)} & \textbf{F1 (\%)} \\
\midrule
\multirow[c]{4}{*}{BERT} & SciTechBaitRo & 89.34 & \textbf{\underline{87.15}} & 87.87 \\
 & SaRoCo & 87.75 & 85.25 & 86.01 \\
 & RoCliCo & \textbf{\underline{90.29}} & 86.76 & \textbf{\underline{88.18}} \\
 & None& 85.98 & 86.24 & 85.13 \\
\cline{1-5}
\multirow[c]{4}{*}{Distilled-BERT} & SciTechBaitRo & 82.10 & \underline{84.94} & 83.20 \\
 & SaRoCo & 83.40 & 84.91 & 83.84 \\
 & RoCliCo & 81.82 & 84.30 & 82.69 \\
 & None & \underline{87.84} & 81.73 & \underline{84.57} \\
\cline{1-5}
\multirow[c]{4}{*}{XLM-RoBERTa} & SciTechBaitRo & 71.88 & 72.95 & 71.47 \\
 & SaRoCo & 73.21 & \underline{73.23} & 72.10 \\
 & RoCliCo & 71.88 & 72.95 & 71.47 \\
 & None & \underline{76.92} & 71.15 & \underline{73.26} \\
\bottomrule
\end{tabular}
}
    \caption{Average model performance across the news topics for the test set. Underline indicates the best score for each model, and bold indicates the best overall scores.}
    \label{tabel_average_test_BERT}
\end{table}

\subsection{Reptile}

\begin{table}[!t]
    \centering
    {
    \small
    \begin{tabular}{lllllll}
\toprule
\textbf{Model} & \textbf{R (\%)} & \textbf{P (\%)} & \textbf{F1 (\%)} \\
\midrule
BERT           &  \textbf{93.96}&  \textbf{89.25}&  \textbf{91.18}\\
Distilled-BERT &           90.44 &           86.83 &           88.25\\
XLM-RoBERTa    &           81.84 &            88.2 &            84.8 \\
\bottomrule
\end{tabular}
}
    \caption{Average model results across topics for the test set when using Reptile.}
    \label{avg_test_Reptile_BERT}
\end{table}

We report in Table \ref{avg_test_Reptile_BERT} the average value of each metric for each model across the news topics in the test set.
The results show that BERT obtained the highest F1-scores, and also obtained the highest scores for most other metrics. During experiments, we observed that XLM-RoBERTa achieved the highest recall for non-satirical content, demonstrating that a multilingual model can adapt to more specific language features and even surpass those trained explicitly for a single language when considering specific metrics. 

\begin{figure}[!t]
\centering
    \includegraphics[width=0.47\textwidth]{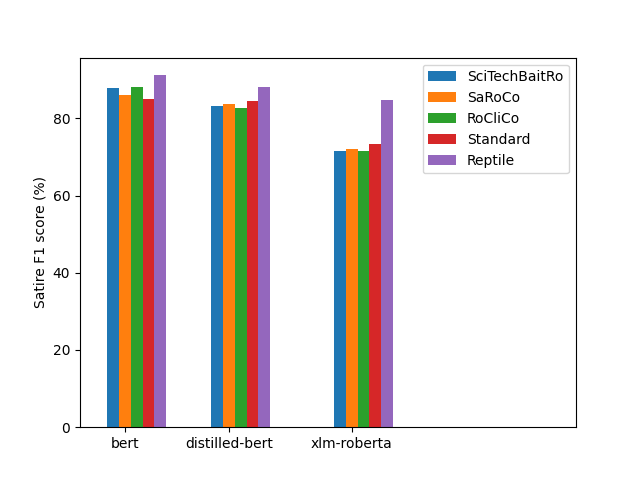}
\caption{Average F1 for the models with and without intermediate tasks compared with Reptile.}
    \label{bar_plot_reptile_vs_all}
\end{figure}

Figure \ref{bar_plot_reptile_vs_all} shows the average F1 satire scores when BERT is fine-tuned with and without intermediate unsupervised transfer learning, and also when the whole model was trained using Reptile. The Reptile algorithm successfully improved the metric of each model, resulting in F1 values that are higher than those achieved when the model underwent intermediate training on only one intermediate task. This finding shows that the ability of "learning to learn" enhances their capacity to learn relevant features.

\subsection{Large Language Models}

\subsubsection{Zero- and Few-shot Learning}

Table \ref{avg_LLM_model_test} shows that RoLlama3 obtained the highest satirical F1-score for satire, few-shot wise, when receiving six examples. A surprising result comes from RoMistral, which underperformed in satire detection in both zero- and 2-shot settings, indicating that the model requires numerous examples to learn the patterns in a satirical headline, as demonstrated at 4- and 6-shot settings.
RoLlama3 achieved the highest F1-score for non-satirical headlines at the 0-shot setting, indicating a potential bias that may stem from the training data. It also exhibited an increasing F1 satire score as the number of examples grew. In the case of RoMistral, apart from the poor start at zero and 2-shot setting, the satire metrics improved at the 4-shot setting and decreased at the 6-shot setting. These behaviors demonstrate that some models can adapt effectively to an increasing number of examples, while others do not.

\begin{table}[ht]
    \centering
    {
    \small
    \begin{tabular}{lllll}
\toprule
\textbf{N-shot} & \textbf{Model} & \textbf{R (\%)} & \textbf{P (\%)} & \textbf{F1 (\%)} \\
\midrule
\multirow[c]{4}{*}{0} & RoLlama2 & 66.83 & 51.24 & 57.95 \\
 & RoLlama3 & \underline{83.03} & \underline{56.85} & \underline{67.49} \\
 & RoGemma & 74.31 & 51.64 & 60.88 \\
 & RoMistral & 21.69 & 49.06 & 30.06 \\
\cline{1-5}
\multirow[c]{4}{*}{2} & RoLlama2 & 48.76 & 55.17 & 51.68 \\
 & RoLlama3 & \textbf{\underline{93.22}} & 58.68 & \underline{72.03} \\
 & RoGemma & 72.96 & 50.78 & 59.87 \\
 & RoMistral & 6.22 & \underline{60.83} & 11.19 \\
\cline{1-5}
\multirow[c]{4}{*}{4} & RoLlama2 & 57.39 & 51.23 & 53.96 \\
 & RoLlama3 & \underline{88.47} & \underline{61.67} & \underline{72.66} \\
 & RoGemma & 59.67 & 52.41 & 55.79 \\
 & RoMistral & 26.41 & 51.90 & 34.91 \\
\cline{1-5}
\multirow[c]{4}{*}{6} & RoLlama2 & 56.36 & 52.07 & 54.02 \\
 & RoLlama3 & \underline{89.45} & \textbf{\underline{62.5}} & \textbf{\underline{73.53}} \\
 & RoGemma & 54.31 & 50.84 & 52.46 \\
 & RoMistral & 25.13 & 51.10 & 33.59 \\
\bottomrule
\end{tabular}
    }
    \caption{Average few-shot metrics on the test set. Underline indicates the best score for each n-shot experiment, and bold indicates the best overall scores.}
    \label{avg_LLM_model_test}
\end{table}

\subsubsection{LoRA Fine-tuning}

Table \ref{avg_LLM_lora_test} reports the average metrics for the test set across the news topics. Comparing with the Table \ref{avg_LLM_model_test}, LoRA significantly improved the results obtained in the few-shot setting for all models except RoGemma. RoLlama2 obtained the highest F1-scores, although slightly less than the ones obtained for BERT in the meta-learning case, as outlined in the Tables \ref{avg_test_Reptile_BERT}. RoLlama3 and RoMistral's performances were similar to the BERT variations in the intermediate task transfer learning setup. 

\begin{table}[!t]
    \centering
    {
    \small
\begin{tabular}{llll}
\toprule
\textbf{Model} & \textbf{R (\%)} & \textbf{P (\%)} & \textbf{F1 (\%)} \\
\midrule
RoGemma & 53.43 & 73.57 & 61.58 \\
RoMistral & 89.31 & 91.45 & 90.25 \\
RoLlama2 & 64.31 & \textbf{97.98} & 75.13 \\
RoLlama3 & \textbf{90.27} & 90.57 & \textbf{90.39} \\
\bottomrule
\end{tabular}

}
    \caption{Average LoRA metrics for test set.}
    \label{avg_LLM_lora_test}
\end{table}

\section{Conclusions}

In this paper, we introduce a dataset compiled from headlines gathered from reputable news websites featuring humorous content inspired by real-life events from Romania.
We evaluated various baselines based on standard machine learning algorithms, BERT, and LLMs. Attempting to select only a specific number of features based on statistical tests, such as ANOVA, Chi-Squared, or entropy-based methods, like Mutual Information, did not yield significant differences. BERT has shown its enhanced capability to capture the overall context, making it a suitable classifier.
Reptile has boosted the performance, highlighting that the models generalize better during a 'learning to learn' from only a few examples from each sampled task in a meta-iteration. LLMs have proved that they struggle with the underlying structure of the satirical content unless directly fine-tuned. The improvement observed by using LoRA fine-tuning suggests that the LLM's ability to classify accurately is limited without prior exposure.

\section*{Limitations}

Our work presents several limitations.

\textbf{Missed Named Entities.} We know that several entities could not be recognized and labeled accurately by Spacy, and as such, we had to label some words manually. We acknowledge that we could not manually label all words not caught by Spacy, which constitutes a limitation.

\textbf{Source of originally satirical content.} The extracted satirical content comes only from a single source, and, as a result, there is a possibility of bias towards a specific kind of satirical content.

\textbf{LLM prompt complexity.} We chose the LLM prompt mainly for its simplicity, and we acknowledge that directly prompting an LLM to provide a verdict by giving as input only a title may not be sufficient, as the model is not offered with other hints that can impact its results.

\textbf{Number of intermediate tasks.} Because there have not been so many related datasets to the one we worked with in the Romanian language, this is considered a limitation for the intermediate unsupervised training of the BERT models.

\section*{Ethical Considerations}

The research was conducted in accordance with the copyright and intellectual property rights, under the European Union directive 2019/790\footnote{\url{https://eur-lex.europa.eu/eli/dir/2019/790/oj/eng}}. As such, the publicly available collected data was used solely for non-commercial research purposes under the provisions of Article 3.
The labeling of named entities was done in compliance with GDPR regulations\footnote{\url{https://eur-lex.europa.eu/eli/reg/2016/679/oj/eng}}. 

\bibliography{custom}

\newpage
\appendix

\section{Prompts for Large Language Models}
\label{app:prompts}

The prompts in the Romanian language are as follows.

\begin{tcolorbox}[title=System]
Ești un bun cunoscător al elementelor care definesc satira. Satira are ca scop ridiculizarea unor comportamente, tocmai de aceea se pot regăsi elemente de absurd, ironie, sarcasm.
\end{tcolorbox}

\begin{tcolorbox}[title=User]
 Vei primi un titlu dintr-o știre și trebuie să spui dacă acesta este satiric, sau nu. 
Vei răspunde numai cu 'da', sau 'nu', fără a mai fi necesare alte explicații.
Observație: pentru a ține cont exclusiv de structura titlului, entitățile au fost ascunse. Nu cunoști articolul și în stabilirea verdictului te vei folosi exclusiv de titlu, fără a apela la cunoștințe externe. Dacă unele comportamente sunt foarte grave, răspunde cu 'nu'.

Titlu:\{title\}
\end{tcolorbox}

The translation into English is as follows:

\begin{tcolorbox}[title=System]
You have a good knowledge of the elements that define the satire. Satire has the goal of ridiculing certain behaviours, which is why you can find elements of absurdity, irony, and sarcasm.
\end{tcolorbox}

\begin{tcolorbox}[title=User]
You will receive a headline, and you need to tell if it is satirical or not. You will only answer with 'yes' or 'no', without requiring further explanations. 
Note: to keep in account only the title's structure, the entities were hidden. You do not know its underlying article, and you will exclusively use the headline, without resorting to external knowledge. If some behaviours are concerning, answer with 'no'.

Title:\{title\}
\end{tcolorbox}

\end{document}